\documentclass{article}

\usepackage[final, nonatbib]{nips_2017}

\usepackage[utf8]{inputenc} 
\usepackage[T1]{fontenc}    
\usepackage{hyperref}       
\usepackage{url}            
\usepackage{booktabs}       
\usepackage{amsfonts}       
\usepackage{nicefrac}       
\usepackage{microtype}      

\usepackage[pdftex]{graphicx}
\usepackage{caption}
\usepackage{subcaption}
\usepackage{mathtools}
\usepackage{color}

\usepackage{hyperref}
\usepackage[nolist]{acronym}
\usepackage{subcaption}
\usepackage{makecell}
\usepackage{multicol}
\DeclareMathOperator{\sgn}{sgn}

\begin{acronym}
\acro{RNN}{recurrent neural network}
\acro{GRU}{gated recurrent unit}
\acro{LSTM}{long short-term memory}
\acro{MFCC}{Mel-frequency cepstral coefficient}
\acro{PEM}{per-epoch noise mixing}
\acro{SNR}{signal-to-noise ratio}
\acro{ASR}{automatic speech recognition}
\acro{CNN}{convolutional neural network}
\acro{DNN}{deep neural network}
\acro{ACCAN}{accordion annealing}
\acro{WSJ}{Wall Street Journal}
\acro{CTC}{Connectionist Temporal Classification}
\acro{WFST}{Weighted Finite State Transducer}
\acro{WER}{word error rate}
\acro{CER}{character error rate}
\acro{LVCSR}{large-vocabulary continuous speech recognition}
\acro{ROI}{range of interest}
\acro{STAN}{sensor transformation attention network}
\acro{SER}{sequence error rate}
\acro{HMM}{Hidden Markov Model}
\acro{LER}{label error rate}
\end{acronym}

\DeclarePairedDelimiter\floor{\lfloor}{\rfloor}

\DeclarePairedDelimiter\abs{\lvert}{\rvert}

\title{Sensor Transformation Attention Networks}

\author{Stefan Braun\thanks{Both authors contributed equally to this work}, Daniel Neil\footnotemark[1], Enea Ceolini, Jithendar Anumula, Shih-Chii Liu \\
  Institute of Neuroinformatics\\
  University of Zurich and ETH Zurich\\
  8057 Zurich, Switzerland \\
  \texttt{brauns@ethz.ch}, \texttt{daniel.l.neil@gmail.com}, \texttt{enea.ceolini@ini.uzh.ch},\\
  \texttt{anumula@ini.uzh.ch}, \texttt{shih@ini.ethz.ch}\\
}

\begin{document}

\maketitle

\begin{abstract}
Recent work on encoder-decoder models for sequence-to-sequence mapping has shown that integrating both temporal and spatial attention mechanisms into neural networks increases the performance of the system substantially. In this work, we report on the application of an attentional signal not on temporal and spatial regions of the input, but instead as a method of switching among inputs themselves.
We evaluate the particular role of attentional switching in the presence of dynamic noise in the sensors, and demonstrate how the attentional signal responds dynamically to changing noise levels in the environment to achieve increased performance on both audio and visual tasks in three commonly-used datasets: TIDIGITS, Wall Street Journal, and GRID.  Moreover, the proposed sensor transformation network architecture naturally introduces a number of advantages that merit exploration, including ease of adding new sensors to existing architectures, attentional interpretability, and increased robustness in a variety of noisy environments not seen during training. Finally, we demonstrate that the sensor selection attention mechanism of a model trained only on the small TIDIGITS dataset can be transferred directly to a pre-existing larger network trained on the Wall Street Journal dataset, maintaining functionality of switching between sensors to yield a dramatic reduction of error in the presence of noise.
\end{abstract}

\section{Introduction and Motivation}

Attentional mechanisms have shown improved performance as part of the encoder-decoder based sequence-to-sequence framework for applications such as image captioning~\cite{xu2015show}, speech recognition~\cite{bahdanau2016end}, and machine translation~\cite{bahdanau2014neural,wu2016google}. Dynamic and shifting attention, for example, on salient attributes within an image aids in image captioning as demonstrated by the state-of-art results on multiple benchmark datasets~\cite{xu2015show}.
Similarly, an attention-based recurrent sequence generator network can replace the \ac{HMM} typically used in a large vocabulary continuous speech recognition systems, allowing an \ac{HMM}-free RNN-based network to be trained for end-to-end speech recognition~\cite{bahdanau2016end}.

While attentional mechanisms have been applied to both spatial and temporal features, this work introduces a \textit{sensor-selection} attention method.  Drawing on inspiration from neuroscience~\cite{desimone1995neural}, we introduce a general \ac{STAN} architecture that supports multi-modal and/or multi-sensor input where each input sensor receives its own attention layer and transformation layers.
It uses an attentional mechanism that can more robustly process data in the presence of noise, allows network reuse, and prevents large increases in parameters as more sensory modalities are added. The output of the attentional signal can itself be interpreted easily from this network architecture.

Furthermore, we also introduce a method of training \ac{STAN} models with random walk noise. First, this enables the model to dynamically focus its attention on the sensors with more informative input or lower noise level. Second, this noise type is designed to help the attention mechanism of the model generalize to noise statistics not seen during training. Finally, we show that the sensor selection attention mechanism of a \ac{STAN} model trained on a smaller dataset can be transferred to a network previously trained on a larger, different dataset. 

Our architecture can be seen as a generalization of many existing network types~\cite{hori2017multimodal,kim2016recurrent, xu2015show}; this work aims to extract general properties of networks types that process temporal sequences with numerous and possibly redundant sensory modalities. The network can be extended easily to multiple sensors because of its inherently modular organization, and therefore is attractive for tasks requiring multi-modal and multi-sensor integration.

\section{Sensor Transformation Attention Network}
\label{sec:STAN}
\begin{figure}[ht]
        \includegraphics[width=\textwidth]{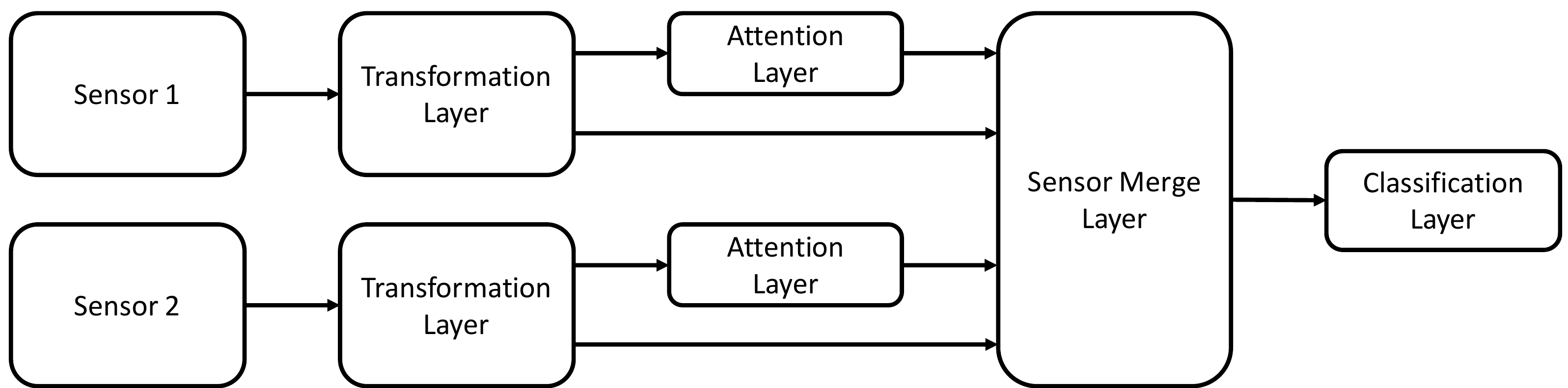}
        \caption{\ac{STAN} model architecture.} 
    \label{fig:stan}
\end{figure}

We introduce the \acp{STAN} in \autoref{fig:stan} as a general network architecture that can be described with five elementary building blocks: (1) input sensors, (2) transformation layers, (3) attention layers, (4) a single sensor merge layer and (5) a stack of classification layers. 

Formally, we introduce a pool of $N$ sensors $s_i$ with $i=1,...,N$ where each sensor provides a feature vector $f_i$. The transformation layers transform the feature vectors $f_i$ to the transformed feature vectors $t_i$. If no transformation layers are used, we maintain the identity $f_i=t_i$. The attention layers compute a scalar attention score $z_i$ for their corresponding input $t_i$. The sensor merge layers compute attention weights $a_i$ by performing a softmax operation over all attention scores $z_i$ (\autoref{softmax}). Each transformed feature vector $t_i$ is then scaled by the corresponding attention weight $a_i$ and merged by an adding operation (\autoref{merge}). The resulting merged transformed feature vector $t_{\text{merged}}$ is then presented to the classification layers for classification.

\noindent\begin{minipage}{.5\linewidth}
\begin{equation}
   a_i(z)=\frac{\exp(z_i)}{\sum\limits_{k=1}^N \exp(z_k)}
   \label{softmax}
\end{equation}
\end{minipage}%
\begin{minipage}{.5\linewidth}
\begin{equation}
t_{\text{merged}} = \sum\limits_{i=1}^N a_i \cdot t_i  \end{equation}
\label{merge}
\end{minipage}

The focus of this work is sequence-to-sequence mapping on time-series in which the attention values are computed on a per-frame basis.  This allows the \ac{STAN} architecture to dynamically adjust to compensate for changes in signal quality due to noise, sensor failure, or informativeness.  As the attentional layer is required to study the dynamics of the input stream to determine signal quality, \acp{GRU}\cite{chung2014empirical} are employed as a natural choice. The transformation layers heavily depend on the input modality, with \acp{GRU} being a good choice for audio features and \acp{CNN}\cite{lecun1998, krizhevsky2012} well adapted for images.


\section{Random Walk Noise Training}
\label{sec:randomwalk}
\begin{figure}[ht]
    \begin{subfigure}[b]{0.31\textwidth}
        \includegraphics[width=\textwidth]{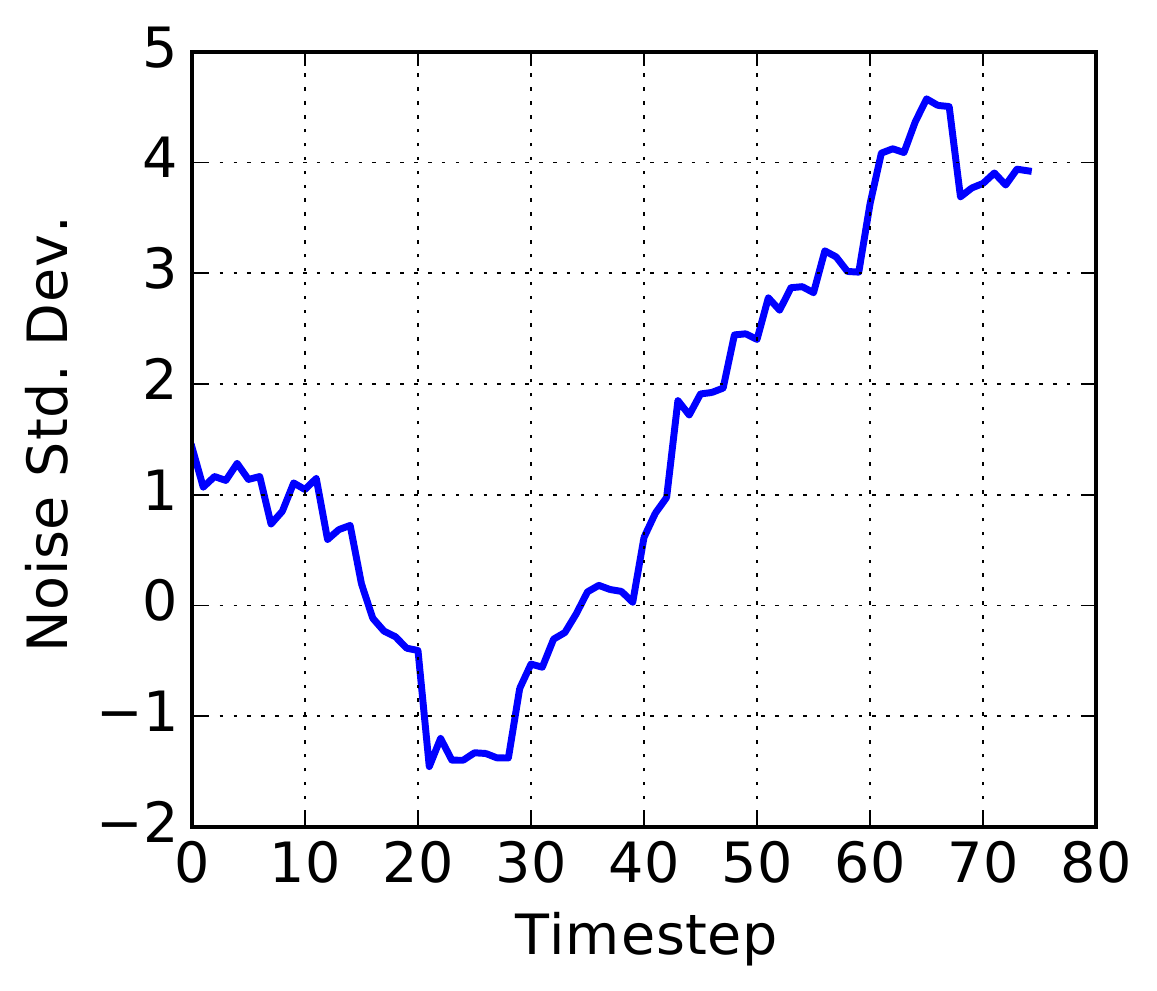}
        \caption{}
        \label{fig:randomwalkcumsum}
    \end{subfigure}
    ~
    \begin{subfigure}[b]{0.31\textwidth}
        \includegraphics[width=\textwidth]{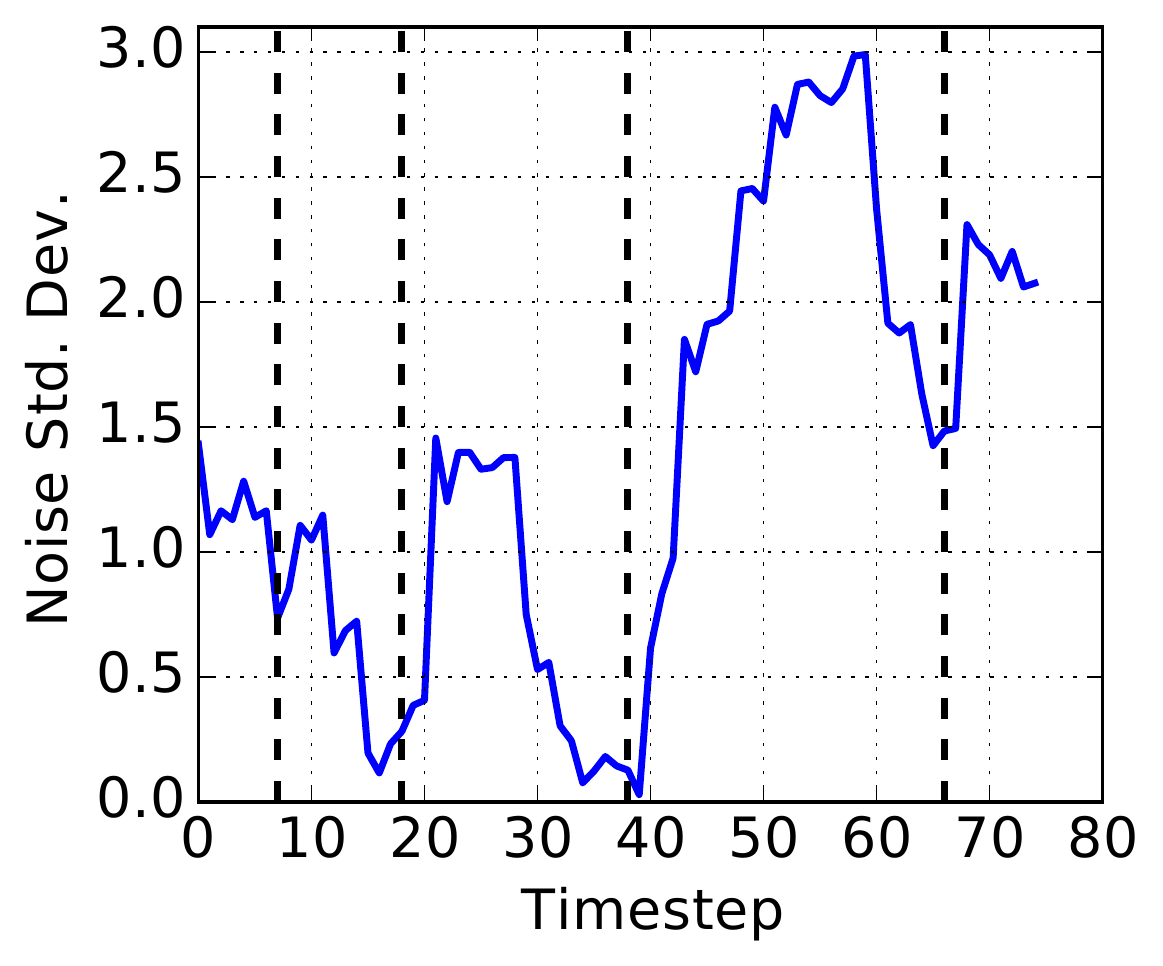}
        \caption{}
        \label{fig:randomwalkfinal}
    \end{subfigure}
    ~
    \begin{subfigure}[b]{0.31\textwidth}
        \includegraphics[width=\textwidth]{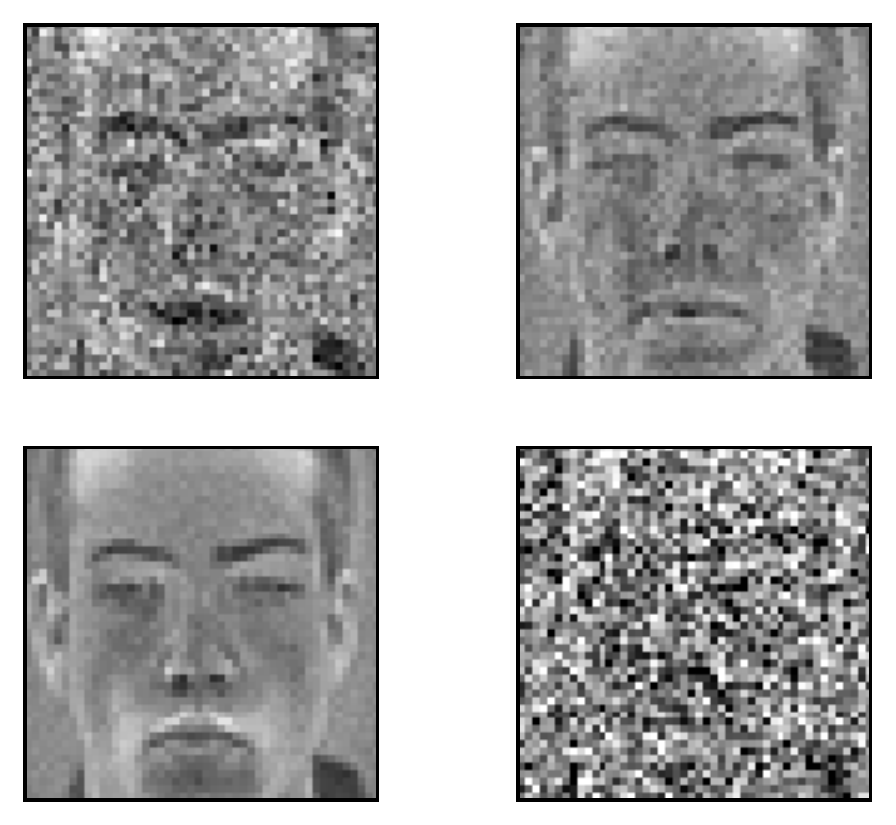}
        \caption{}
        \label{fig:randomwalkex}
    \end{subfigure}    
    \caption{Depiction of the random walk noise added during training.  In \textbf{(a)}, the cumulative sum of a sequence of random variables forms a random walk.  In \textbf{(b)}, the random walk becomes bounded after applying the reflection operator $\phi$ in Eq.~\ref{eq:reflection}.  On the four sub-panels in \textbf{(c)}, a visualization of the noise drawn at each time point.  Each panel depicts a video frame from the GRID corpus, zero mean and unit variance normalized, mixed with a Gaussian noise source whose standard deviation corresponds to a vertical dotted line in \textbf{(b)}.}
    \label{fig:randomwalk}
\end{figure}

To enable the network model to be robust against a wide variety of noise types, we introduce the random walk noise model below.  The noise model aims to have an approximately uniform coverage of noise level over a range $[0, \sigma_{max})$ and no settle-in time that could introduce a sequence length dependence on the noise.  The standard deviation of the noise $\sigma$ for an input sequence of $t$ timesteps can be calculated thusly:

\begin{align}
\begin{split}
\sigma(t) = \phi\left(\underbrace{\sigma_0 + \sum\nolimits_{i=1}^{t} \sgn(s_i) n_i}_{a},\quad \sigma_{max}\right),\\
\sigma_0 \sim \mathcal{U}(0, \sigma_{max}/2), \quad s_i \sim \mathcal{U}(-1,1) \quad n_i \sim \Gamma(k, \theta) 
\label{eq:random_walk}
\end{split}
\end{align}

with $\sigma_0$ drawn uniformly over the range $[0, \sigma_{max}/2)$ and $n_i$ drawn from a gamma distribution with shape $k$ and scale $\theta$. The signum function extracts positive and negative signs from $s_i$ with equal probability. A parameter search during our experiments yielded $\sigma_{max} = 3$, $k=0.8$ and $\theta=0.2$ as an appropriate set of parameters. We define the reflection function $\phi(a, \sigma_{max})$ as
\begin{align}
    \phi(a, \sigma_{max}) = \sigma_{max} - \abs[\Big]{ \bmod (a, 2\sigma_{max}) - \sigma_{max}}
    \label{eq:reflection}
\end{align}
where $\bmod(a, 2\sigma_{max}) = a-2\sigma_{max}\floor{a / 2\sigma_{max}}$ maintains the values within the desired range $(0, 2\sigma_{max})$ and the subsequent shift and magnitude operations map the values to the range $[0, \sigma_{max})$ while avoiding discontinuities.  Finally the input data $x$ at feature index $
k$ and time index $t$ is mixed with the noise sampled from a normal distribution as follows:
\begin{align}
    x_{k,t} = x_{k,t} + n_{k, t}, \qquad
    n_{k, t} \sim \mathcal{N}(0, \sigma^2(t))
\end{align}
The reflection function $\phi(a, \sigma_{max})$ performs similarly to the $\bmod$ operator, but at the edges, produces a continuous reflection about the edge instead of a discontinuous wrap.  Therefore, this forms a constrained random walk, limited by $\sigma_{max}$, which will become the standard deviation of normally distributed random noise added to the input $x$ at feature index $k$ and time point $t$. This noise model generates sequences that provide a useful training ground to tune the attention mechanism of \ac{STAN} models, as the noise level varies over time and allows periods of low noise (high attention desired) and high noise (low attention desired).  
 An example for video frames is shown in \autoref{fig:randomwalk}. 

\section{Experiments}
\label{sec:expt}

This section presents experiments that show the performance of \acp{STAN} in environments with dynamically changing noise levels and tested on three commonly-used datasets, namely TIDIGITS, Wall Street Journal and GRID.

\subsection{Noise Experiments}
\label{sec:tidigits}

\paragraph{Dataset}

  \begin{table}[t]
   \caption{Models used for evaluation on TIDIGITS.}
   \label{tab:tidigits_models}
   \centering
   \resizebox{\textwidth}{!}{%
  \begin{tabular}{lcccccc}
        \toprule
        Name & Architecture & Sensors & \makecell{Transformation \\Layers} & \makecell{Attention \\Layers} &\makecell{Classification \\Layers} & Parameters \\
        \midrule
        Single Audio & Baseline & 1 Audio & Identity & None & (150,100) GRU & 162262 \\
        Double Audio STAN & STAN & 2 Audio & Identity & (20) GRU & (150,100) GRU & 169544 \\
        Triple Audio STAN & STAN & 3 Audio & Identity & (20) GRU & (150,100) GRU & 173185 \\
        Double Audio Concat & Concatenation & 2 Audio & Identity &  None & (150,100) GRU & 179812 \\
        Triple Audio Concat & Concatenation & 3 Audio & Identity & None & (150,100) GRU & 197362 \\  
   \end{tabular}}
 \end{table}

The \textit{TIDIGITS} dataset~\cite{leonard1993tidigits} was used as an initial evaluation task to demonstrate the response of the attentional signal to different levels of noise in multiple sensors. 
The dataset consists of 11 spoken digits (`oh', `zero' and `1' to `9') in sequences of 1 to 7 digits in length, e.g `1-3-7' or `5-4-9-9-8'. The dataset is partitioned into a training set of 8623 samples and a test set of 8700 samples. 
The raw audio data was converted to 39-dimensional \acp{MFCC} \cite{davis1980comparison} features (12 Mel spaced filter banks, energy term, first and second order delta features). A frame size of 25ms and a frame shift of 10ms were applied during feature extraction. The features are zero-mean and unit-variance normalized on the whole dataset. The \ac{SER} is used as a performance metric. 
It is defined as the number of entirely correct sequence transcriptions $C$ over the number of all sequences $A$: SER [\%] = $\frac{C}{A}$. 

\paragraph{Models}
A total of five models were evaluated, with a summary given in \autoref{tab:tidigits_models}. The classification stage is the same two-layer unidirectional (150,100) \ac{GRU} network followed by an affine transform to 12 classes (blank label + vocabulary) for all models. The baseline model consists of a single audio sensor that is directly connected to the classification stage. Two models make use of the \ac{STAN} architecture with two or three sensors. The attention modules consist of (20) \acp{GRU} and their output is converted to one scalar attention score per frame by an affine transform. In order to evaluate the potential benefit of \ac{STAN} architectures, they are compared against two simpler sensor concatenation models. Those two have either two or three sensors, whose input is concatenated and presented directly to the classification network. In these models, the transformation layers were simply the identity function as the task is sufficiently simple to not require their use.  The number of parameters is approximately equal for all models and depends only on the number of input sensors.

\paragraph{Training}
The connected digit sequences allow for a sequence-to-sequence mapping task. In order to automatically learn the alignments between speech frames and label sequences, the \ac{CTC} \cite{graves2006connectionist} objective was adopted. All models were trained with the ADAM optimizer \cite{kingma2014adam} for a maximum of 100 epochs, with early stopping preventing overfitting. Every model was trained on a noisy training set corrupted by random walk noise as detailed in \autoref{sec:randomwalk}. Each sensor received a unique, independently drawn noise signal per training sample. The noise level of the random walks varied between $[0,3\sigma]$. 

\paragraph{Results}

\begin{figure} 
\centering
\includegraphics[width=\textwidth]{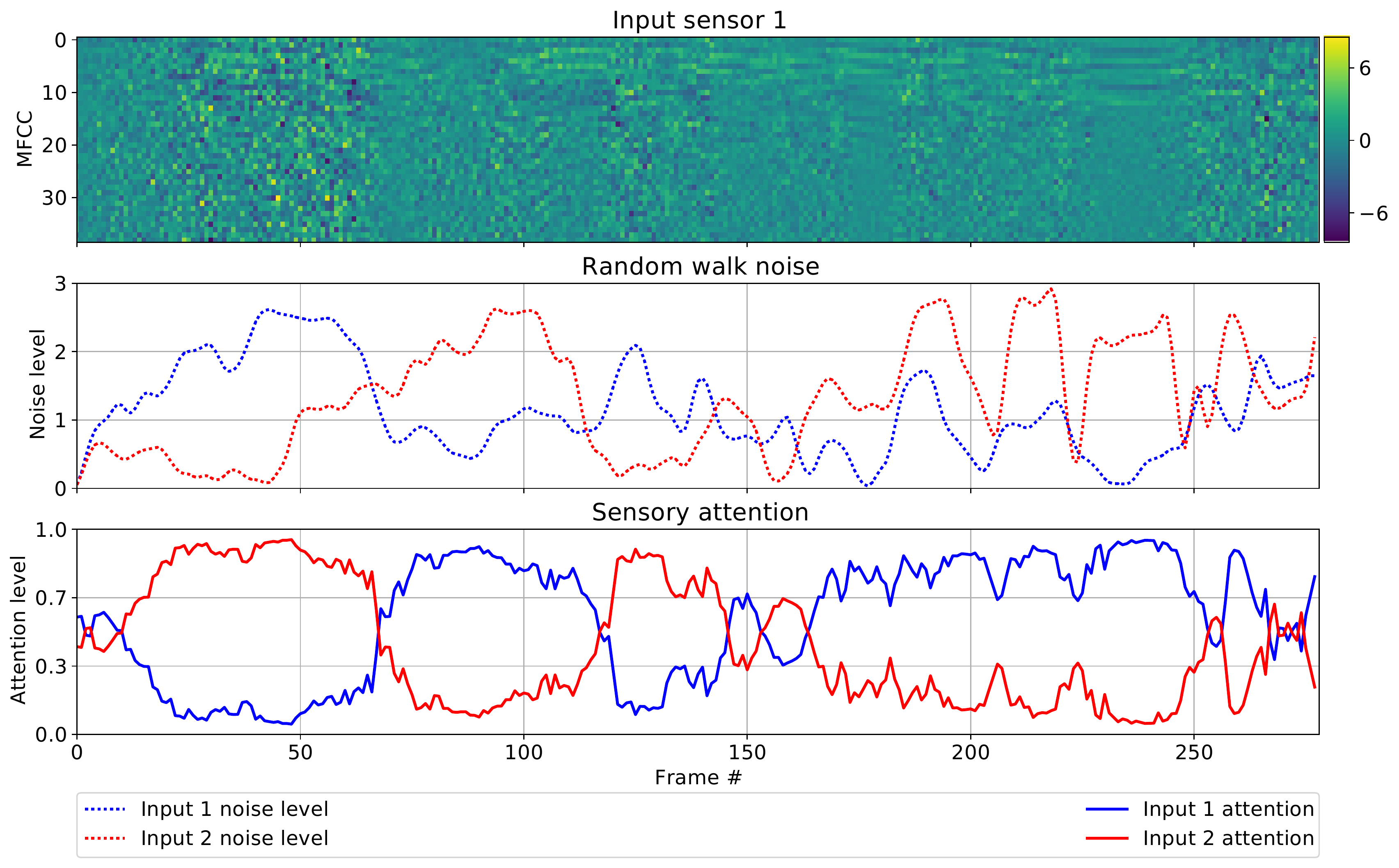}
\caption{Attention response to random walk noise conditions of a double audio \ac{STAN} model trained on TIDIGITS. The figure shows the MFCC features with noise of sensor 1 (top), the noise levels applied to both sensors (middle) and the attention values for both sensor (bottom). The \ac{STAN} model shows the desired negative correlation between noise level and attention for sensors.}
\label{rwn_tidigits}
\end{figure}

\begin{figure}
\centering
\begin{subfigure}[b]{0.49\textwidth}
        \includegraphics[width=\textwidth]{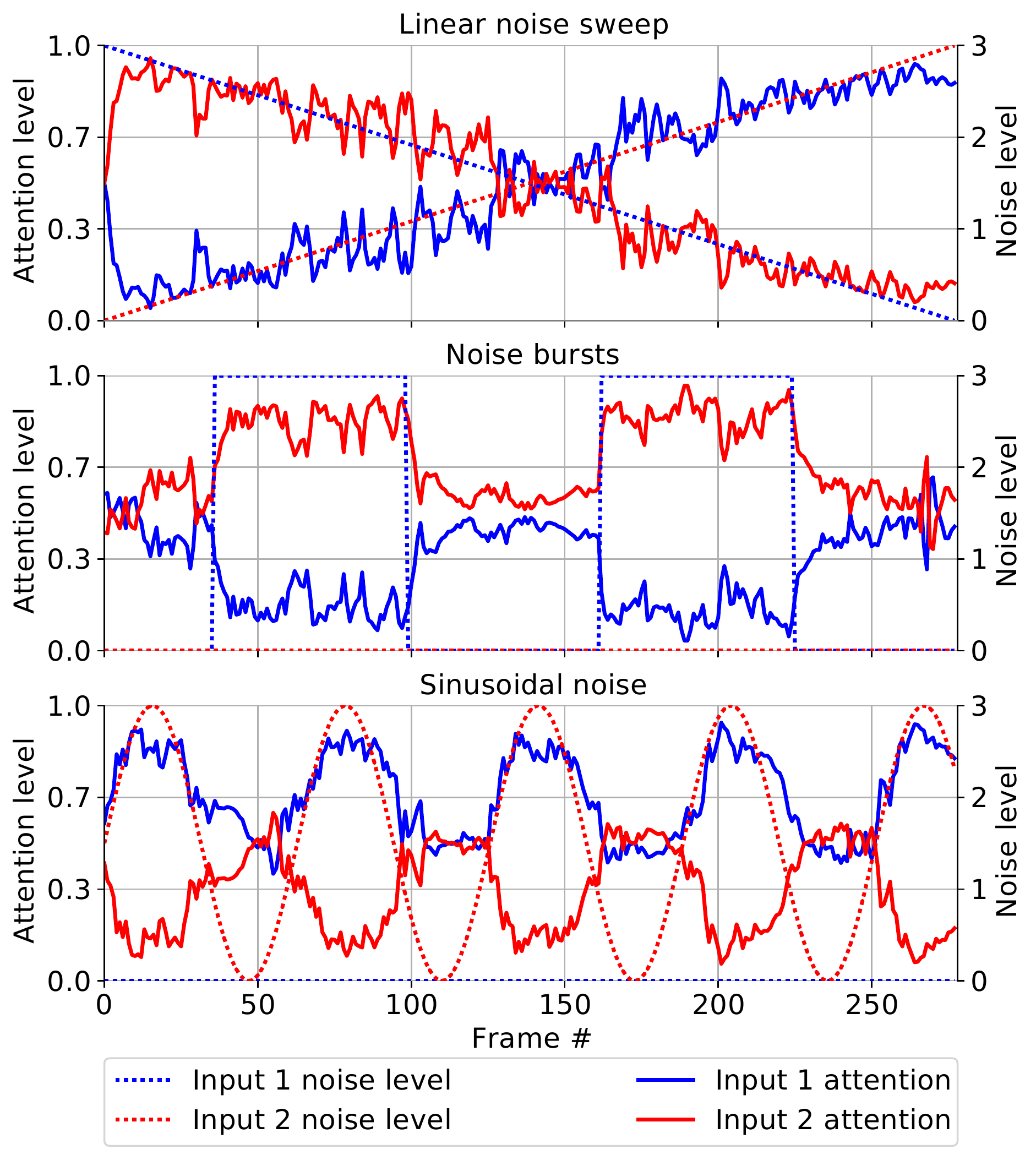}
        \label{mnt_tidigits}
    \end{subfigure}
    ~
\begin{subfigure}[b]{0.49\textwidth}
    \includegraphics[width=\textwidth]{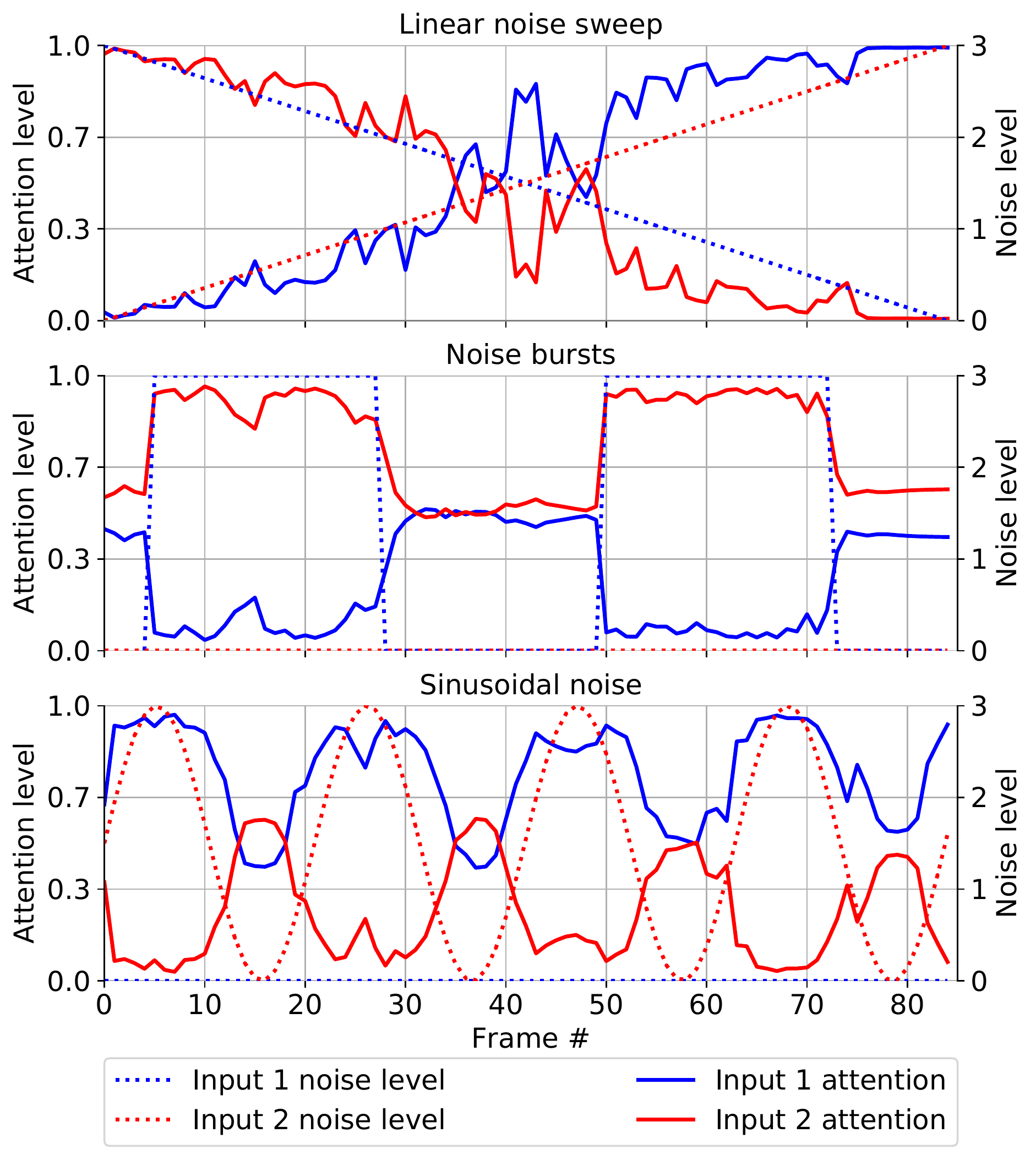}
    \label{mnt_grid}
\end{subfigure}
\caption{Attention response to various noise conditions of a double sensor \ac{STAN} model trained on audio (TIDIGITS, left side) and video (GRID, right side).  Three noise responses are shown: linear noise sweeps on both sensors (top), noise bursts on sensor 1 (middle) and sinusoidal noise on sensor 2 (bottom). Though these noise conditions were not seen during training, both \ac{STAN} models show the desired negative correlation between noise level and sensor attention.}
\label{noise_types}
\end{figure}

Two key results emerge from this initial experiment: first, the attention mechanism generalizes across a variety of noise types; and, secondly, \ac{STAN} models outperform, in error rate, merely concatenating input feature together.
To demonstrate, \autoref{rwn_tidigits} shows the attention response of a Double Audio \ac{STAN} model with two audio sensors in response to random walk noise. A sample from the test set was corrupted by random walk with a noise level in the range $[0,3\sigma]$. The model shows the desired negative correlation between noise level and attention: when the noise level for a sensor goes up, the attention paid to the same sensor goes down. As the noise levels interleave over time, the attention mechanism is able to switch between sensors by a delay of 1-5 frames. Furthermore, without additional training, the same model is evaluated against novel noise types in \autoref{noise_types}. The attention modules successfully focus their attention on the sensor with the lowest noise level under a variety of noise conditions. In situations where the noise level of both sensors is low, such as seen in the noise burst or sinusoidal noise cases, the attention settles in an equilibrium between both sensors.

To determine whether the attention across sensors actually improves performance, the \ac{STAN} models are evaluated against a baseline single sensor model and concatenation models under both clean and noisy conditions.  With the clean testset, all available sensors are presented the same clean signal. With the noisy testset, each sensors data is corrupted by unique random walk noise with a standard deviation in the range $[0,3\sigma]$. The results are reported in \autoref{bars}(a). All models achieve comparably low \ac{SER} around $3\%$ on the clean test set, despite training on noisy conditions, implying the \ac{STAN} architecture does not have negative implications for clean signals. On the noisy test set, the \ac{STAN} models with two and three sensors perform best. The \ac{STAN} models lower the \ac{SER} by 66.8\% (single vs. double sensors) and 75\% (single vs. triple sensors) through additional inputs.   The \ac{STAN} models dramatically outperform the concatenation models with an equivalent number of sensors, achieving around half the \ac{SER}, suggesting concatenation models had difficulties in prioritizing signal sources with lower noise levels.

\subsection{Transfer of Sensor Selection Attention Mechanism to WSJ Corpus}
\paragraph{Dataset} This experiment demonstrates the possibility to train a \ac{STAN} model on a small dataset (TIDIGITS) and transfer the sensor selection attention mechanism to a non-\ac{STAN} model that was independently trained on a much bigger dataset (\ac{WSJ}). The \ac{WSJ} database consists of read speech from the Wall Street Journal.  Following standard methodology~\cite{miao2015eesen}, the 81 hour subset `si284' was used as training set (37416 sentences), the subset `dev93' as development set (513 sentences) and the `eval92' (330 sentences) subset was used as test set.

For both datasets, the raw audio data is converted to 123 dimensional filterbank features (40 filterbanks, 1 energy term and first and second order delta features). During feature extraction, the same frame size of 25ms and frame shift of 10ms were used on both datasets, resulting in longer frame sequences on \ac{WSJ}. The features were generated by pre-processing routines from EESEN \cite{miao2015eesen}. Each feature dimension is zero-mean and unit-variance normalized.

\paragraph{Models}
The TIDIGITS-\ac{STAN} model uses two audio sensors that provide filterbank features, identity transformation layers and 60 \acp{GRU} per attention layer followed by an learned outer product transform to a single attention score per frame. The classification stage on top of the sensor merge layer is built of a unidirectional two-layer (150, 100) \ac{GRU} network followed by an affine transform to 12 classes. The TIDIGITS-\ac{STAN} model uses 267k parameters, with the classification stage accounting for 200k parameters (75\%).

The \ac{WSJ}-baseline model is a non \ac{STAN} model that is trained independently and consists of 8.5M parameters. It is built of 4 layers of bidirectional \acp{LSTM} \cite{hochreiter1997long} units with 320 units in each direction, followed by an affine transformation that maps the last layers output to the 59 output labels (blank label + characters). The \ac{WSJ}-baseline model maps filterbank feature sequences to character sequences. Similar architectures can be found in literature \cite{miao2015eesen}. We build a WSJ-\ac{STAN} model by the following recipe: first, the TIDIGITS-\ac{STAN} model is trained. Second, we train the \ac{WSJ}-baseline model. Third, we replace the classification stage of our TIDIGITS-\ac{STAN} model with the \ac{WSJ}-baseline model. The result of this replacement process is the \ac{WSJ}-\ac{STAN} model, on which no retraining is performed. The standard performance metric on \ac{WSJ} is the \ac{WER}, which is defined as following: \ac{WER} [\%] = $\frac{S+D+I}{N}$, with word level substitutions $S$, deletions $D$, insertions $I$ and the number of words in the reference $N$. 
\paragraph{Training}
Both TIDIGITS and \ac{WSJ} allow for a sequence-to-sequence mapping task. In order to automatically learn the alignments between speech frames and label sequences, the \ac{CTC} \cite{graves2006connectionist} objective was adopted. The models were trained with the ADAM optimizer \cite{kingma2014adam} for a maximum of 100 epochs, with early stopping preventing overfitting. The TIDIGITS-\ac{STAN} model was trained on TIDIGITS training set which was corrupted by random walk noise as detailed in \autoref{sec:randomwalk}. Each sensor received a unique, independently drawn noise signal per training sample again in the range $[0,3\sigma]$. The \ac{WSJ}-baseline model was trained on the \ac{WSJ} `train-si84' training set, which contained clean speech only. 

\paragraph{Results}
The \ac{WSJ}-baseline model and the \ac{WSJ}-\ac{STAN} model are evaluated on the `eval92' test set from the \ac{WSJ} corpus. In \autoref{tab:wsj} we report the \ac{WER} after decoding the network output with a trigram language model based on \acp{WFST} \cite{mohri2008speech}, see \cite{miao2015eesen} for details. For the clean speech test, the same clean signal is used as input for both sensors of the \ac{WSJ}-\ac{STAN} model, it should thus be equivalent to the baseline model in the clean test case. This is confirmed as the both the \ac{WSJ}-baseline and the \ac{WSJ}-\ac{STAN} models achieve the same 8.4\% \ac{WER} on clean speech, which is close to state-of-the art with comparable setups \cite{miao2015eesen}. In the noisy tests, the input features are overlaid with random walk noise with a noise level of up to $3\sigma$. There, the \ac{WSJ}-\ac{STAN} model achieves 26.1\% \ac{WER}, while the \ac{WSJ}-baseline model has over double the error rate at 53.5\% \ac{WER}. This result demonstrates clearly that the \ac{STAN} architecture can generalize its sensor selection attention mechanism to different datasets and different classification stages. It is notable that even though the average number of frames per sample in the \ac{WSJ} `eval92' test set (760) is approximately 4.6X larger than that of the TIDIGITS test set (175), the attention mechanism still functions well.  

\begin{table}[t]
   \caption{Evaluation results on the \ac{WSJ} corpus: \ac{WER} in [\%] after decoding}
   \label{tab:wsj}
   \centering
      \resizebox{0.4\textwidth}{!}{%
  \begin{tabular}{lcccccc}
        \toprule
        Model &  \ac{WSJ}-baseline & \ac{WSJ}-\ac{STAN}\\
        \midrule
        Clean test set & 8.4 & 8.4 \\
        Noisy test set & 53.5 & 26.1
        
   \end{tabular}}
 \end{table}

\subsection{Correct Fusion from Multiple Sensors on Grid}
\label{sec:grid}
\paragraph{Dataset} The \textit{GRID}~\cite{cooke2006audio} corpus is used for perceptual studies of
speech processing. There are 1000 sentences spoken by each of the 34 speakers. The GRID word vocabulary contains four commands (`bin', `lay', `place', `set'), four colors (`blue', `green', `red', `white'), four prepositions (`at', `by', `in', `with'), 25 letters (`A'-`Z' except `W'), ten digits (`0'-`9') and four adverbs (`again', `now', `please', `soon'), resulting in 51 classes. There are 24339 training samples and 2661 test samples, consisting of both audio and video data. The raw audio data was converted to 39-dimensional \acp{MFCC} features (12 Mel spaced filter banks, energy term, first and second order delta features). During feature extraction, a frame size of 60ms and a frame shift of 40ms were applied to match the frame rate. The video frames are converted to grey level frames. Both audio and video data are normalized to zero-mean and unit-variance on the whole dataset. As for TIDIGITS, the \ac{SER} is used as a performance metric.

\paragraph{Training}
The video and audio sequences of the GRID database allow for a sequence-to-sequence mapping task. In order to automatically learn the alignments between speech frames, video frames and label sequences, the \ac{CTC} \cite{graves2006connectionist} objective was adopted. The output labels consisted of 52 classes (vocabulary size + blank label). All models were trained with the ADAM optimizer \cite{kingma2014adam} for a maximum of 100 epochs, with early stopping preventing overfitting. Every model was trained on a noisy training set corrupted by random walk noise as detailed in \autoref{sec:randomwalk}. Each sensor received a unique, independently drawn noise signal per training sample. The noise level of the random walks varied in the range $[0,3\sigma]$.

 \begin{table}[t]
   \caption{Models used for evaluation on GRID.}
   \label{tab:grid_models}
   \centering
   \small
   \resizebox{\textwidth}{!}{%
  \begin{tabular}{lcccccc}
        \toprule
        Name & Architecture & Sensors & \makecell{Transformation \\Layers} & \makecell{Attention \\Layers} &\makecell{Classification \\Layers} & Parameters \\        
        \midrule
        Single Audio & Baseline & 1 Audio & (50) Dense & None & (200,200) BI-GRU & 1030012 \\
        Double Audio STAN & STAN & 2 Audio & (50) Dense &  (20) GRU & (200,200) BI-GRU & 1056654 \\
        Triple Audio STAN & STAN & 3 Audio & (50) Dense & (20) GRU & (200,200) BI-GRU & 1062955 \\
        Double Audio Concat & Concatenation & 2 Audio & (50) Dense & None & (200,200) BI-GRU &  1108052 \\
        Triple Audio Concat & Concatenation & 3 Audio & (50) Dense & None & (200,200) BI-GRU &1170052 \\
        \midrule
        Single Video & Baseline & 1 Video & CNN &  None & (200,200) BI-GRU &1061126 \\
        Double Video STAN & STAN & \makecell{2 Video} & \makecell{CNN} & (150) GRU & (200,200) BI-GRU & 1087562 \\
   \end{tabular}}
 \end{table}
 
\paragraph{Models}
A total of seven models were evaluated: five models that use audio input only and two model that use video input only. A summary is given in \autoref{tab:grid_models}.
All models use a two-layer bidirectional \ac{GRU} network with (200, 200) units in each direction followed by an affine transform to 52 classes (blank label + vocabulary) for the classification stage. 
The audio-only models consist of a baseline single sensor model, two \ac{STAN} models with either two or three sensors and two concatenation models with two or three sensors. Every audio sensor makes use of a (50) unit non-flattening dense layer with a $tanh$ non linearity for feature transformation. For the \ac{STAN} models, the attention layers operate on the transformed features and use 20 \acp{GRU} per sensor. Their output is converted to one scalar attention score per frame by an affine transform. The video-only models use a \ac{CNN} for feature transformation: three convolutional layers of 5x5x8 (5x5 filter size, 8 features), each followed by a 2x2 max pooling layer. The output is flattened and presented to classification layer. The double video \ac{STAN} model uses attention layers with 150 \acp{GRU} per sensor.

\begin{figure}
\centering
  \includegraphics[width=\textwidth]{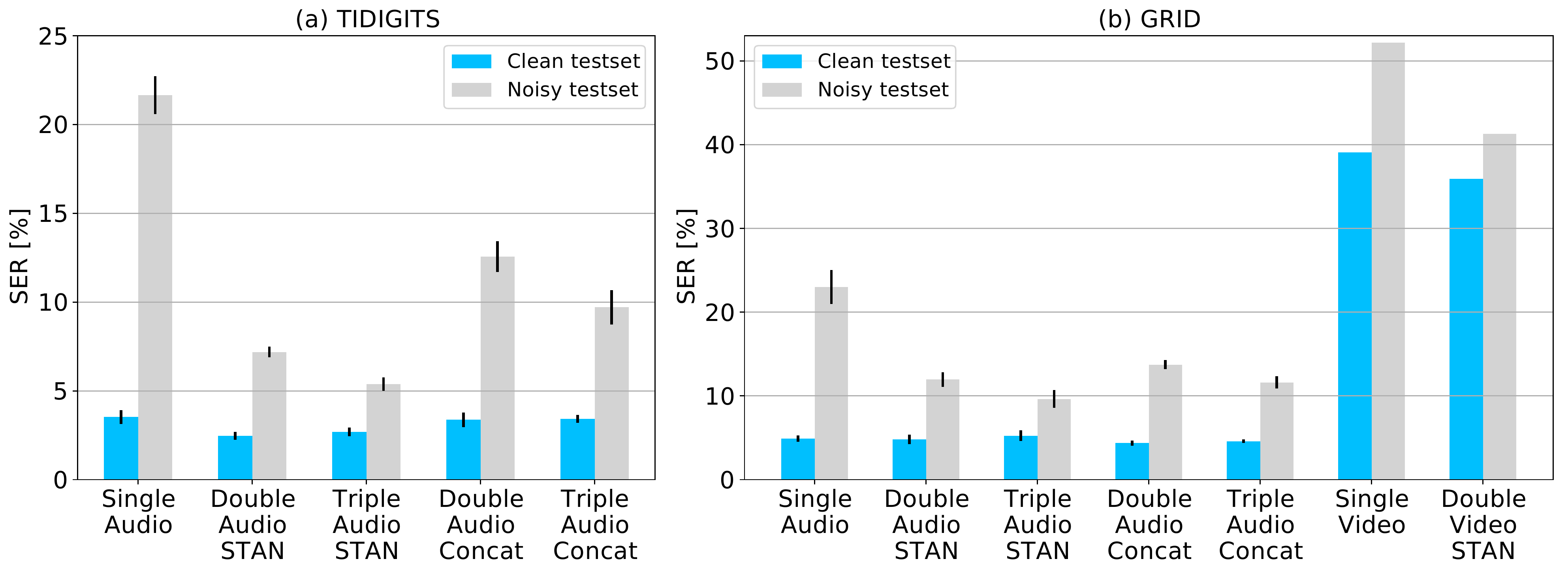}
  \caption{\ac{SER} scores on (a) TIDIGITS and (b) GRID datasets for clean and noisy testsets. The audio scores represent the mean of 5 weight initializations, while the video scores are based on a single weight initialization. The \ac{SER} is a tough error measure that penalizes the slightest error in a sequence transcription. Our GRID video models achieve a 90.5\% (Single Video) and 91.5\% (Double Video \ac{STAN}) label-wise accuracy on clean video. These results compare with state-of-the-art \cite{assael2016lipnet} without data augmentation or architecture specialization.}
  \label{bars}

\end{figure}

\paragraph{Results}
The seven previously described models are compared by their \ac{SER} on the GRID test set. The testing is carried out on a clean variant and a noise corrupted variant of the test set. With the clean test set, all sensors of the same modality are presented with the same clean signal. With the noisy test set, each sensor's data is corrupted by unique random walk noise with a noise level in the range $[0,3\sigma]$. The results are reported in \autoref{bars}(b). 
All of the audio-only models achieve comparably low \ac{SER} around $5\%$ on the clean test set, even though they were trained on noisy conditions, echoing the results found for TIDIGITS.
On the noisy test set, the audio \ac{STAN} models are able to outperform their concatenation counterparts by 13\% (two sensors) and 17\% (three sensors). Adding more sensors to the \ac{STAN} models relatively lowers the \ac{SER} by 48\% (single vs. double audio sensors) and 58\% (single vs. triple audio sensors). 

The video-only baseline model achieves a performs worse than the single audio-only model on both clean and noisy test conditions. However, the \ac{STAN} model is still able to improve the \ac{SER} scores. The sensor selection attention mechanism also works on video data, as depicted in \autoref{noise_types}.  In order to successfully train the \ac{STAN} model for video data, the attentional and transformation layer weights are shared across both sensors.



\section{Discussion}
\label{sec:discussion}
The Sensor Transformation Attention Network architecture has a number of interesting advantages that merit further analysis.  By equipping each sensory modality with a mechanism for distinguishing meaningful features of interest, networks can learn how to select, transform, and interpret their sensory stimuli.  First, and by design, \acp{STAN} exhibit remarkable robustness to noise sources. By challenging these networks during training with dynamic and persistent noise sources, the networks learn to rapidly isolate modalities corrupted by noise sources.
We further show that this results in even better performance than maintaining the full high-dimensional space created by concatenating all inputs together. The sensor selection attention mechanism shows remarkable generalization properties: we demonstrated that it can be trained on a small dataset and be reused on a much bigger dataset that demands a much more powerful classification stage.

This architecture also permits investigation of common latent representations reused between sensors and modalities.  Similar to the ``interlingua'' mentioned in machine translation work~\cite{wu2016google}, perhaps the transformation layer allows sensors to map their inputs to a common, fused semantic space that simplifies classification. 
As has been pointed out in a variety of other studies, attention is a powerful mechanism to aid in network interpretability~\cite{xu2015show}.  Here, attention is used to select among sensors, but this could easily be extended to e.g., foveated representations of the environment, allowing very precise analysis of determining the causality of neural network decisions.  Future work could also investigate the inherent informativeness of the attentional signal itself to determine e.g., the direction of approach of an ambulance in a multi-audio setup.




\small 

\bibliographystyle{abbrv}

\bibliography{main}

\end{document}